\documentclass[10pt,twocolumn,letterpaper]{article}

\usepackage{cvpr}              % To produce the CAMERA-READY version
\usepackage{array,multirow,graphicx}
\usepackage{graphicx}
\usepackage{amsmath}
\usepackage{amssymb}
\usepackage{booktabs}
\usepackage{bm}
\usepackage{breqn}
\usepackage{tabularx}
\usepackage[dvipsnames]{xcolor}
\usepackage{booktabs,caption}
\usepackage[bottom]{footmisc}
\usepackage[flushleft]{threeparttable}
\def\bs{\bm}

\makeatletter
\def\thickhline{%
  \noalign{\ifnum0=`}\fi\hrule \@height \thickarrayrulewidth \futurelet
  \reserved@a\@xthickhline}
\def\@xthickhline{\ifx\reserved@a\thickhline
              \vskip\doublerulesep
              \vskip-\thickarrayrulewidth
             \fi
      \ifnum0=`{\fi}}
\makeatother

\newlength{\thickarrayrulewidth}
\newcommand{\RN}[1]{%
  \textup{\uppercase\expandafter{\romannumeral#1}}%
}

\newcolumntype{Y}{>{\centering\arraybackslash}X}

\definecolor{Magenta}{RGB}{255, 0, 255}
\newcommand\boldyellow[1]{\textcolor{yellow}{\textbf{#1}}}
\newcommand\boldmagenta[1]{\textcolor{Magenta}{\textbf{#1}}}

\usepackage[pagebackref,breaklinks,colorlinks]{hyperref}

\usepackage[capitalize]{cleveref}
\crefname{section}{Sec.}{Secs.}
\Crefname{section}{Section}{Sections}
\Crefname{table}{Table}{Tables}
\crefname{table}{Tab.}{Tabs.}

\def\cvprPaperID{5940} % *** Enter the CVPR Paper ID here
\def\confName{CVPR}
\def\confYear{2022}

\begin{document}

\title{ChiTransformer: Towards Reliable Stereo from Cues}

\author{Qing Su\\
Georgia State University\\
{\tt\small qsu3@gsu.edu}
\and
Shihao Ji\\
Georgia State University\\
{\tt\small sji@gsu.edu}
}
\maketitle

%%%%%%%%% ABSTRACT
\begin{abstract}
Current stereo matching techniques are challenged by restricted searching space, occluded regions and sheer size. While single image depth estimation is spared from these challenges and can achieve satisfactory results with the extracted monocular cues, the lack of stereoscopic relationship renders the monocular prediction less reliable on its own especially in highly dynamic or cluttered environments. To address these issues in both scenarios, we present an optic-chiasm-inspired self-supervised binocular depth estimation method, wherein a vision transformer (ViT) with gated positional cross-attention (GPCA) layers is designed to enable feature-sensitive pattern retrieval between views, while retaining the extensive context information aggregated through self-attentions. Monocular cues from a single view are thereafter conditionally rectified by a blending layer with the retrieved pattern pairs. This crossover design is biologically analogous to the \textit{optic-chasma} structure in human visual system and hence the name, ChiTransformer. Our experiments show that this architecture yields substantial improvements over state-of-the-art self-supervised stereo approaches by 11\%, and can be used on both rectilinear and non-rectilinear (e.g., fisheye) images. The project is available at \footnotesize{\url{https://github.com/ISL-CV/ChiTransformer}}.
\end{abstract}

%It leverages strengths of both monocular and binocular approaches. 

% end-to-end 
% encoder-decoder-structured

%%%%%%%%% BODY TEXT
\begin{figure}[ht]
\vspace{10mm}
\begin{minipage}{\columnwidth}
    \setlength{\tabcolsep}{0pt}
    \renewcommand{\arraystretch}{0.1}
\begin{center}
\scriptsize
\begin{tabular}{c@{}c@{}}

\rotatebox[origin=l]{90}{\makebox[0.25\columnwidth]{\scalebox{1}{ (Left) input}}}
&\includegraphics[width = 0.95\columnwidth]{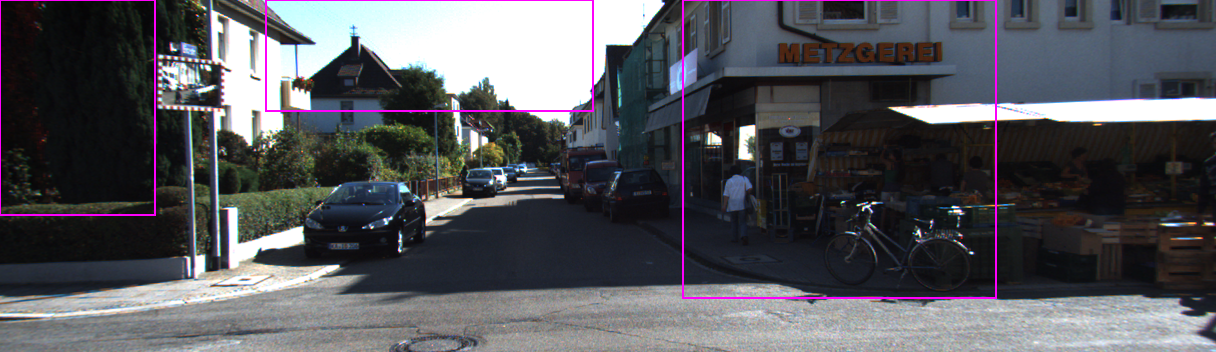}\\

\rotatebox[origin=l]{90}{\makebox[0.25\columnwidth]{\scalebox{1}{ChiTransformer}}}
& \includegraphics[width = 0.95\columnwidth]{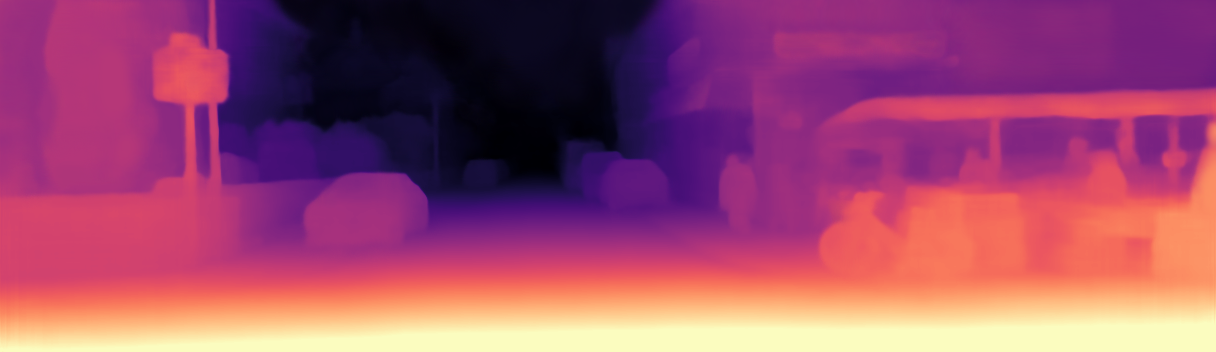}\\

\rotatebox[origin=l]{90}{\makebox[0.25\columnwidth]{\scalebox{1}{DPT \cite{ranftl2021vision}}}}
&\includegraphics[width = 0.95\columnwidth]{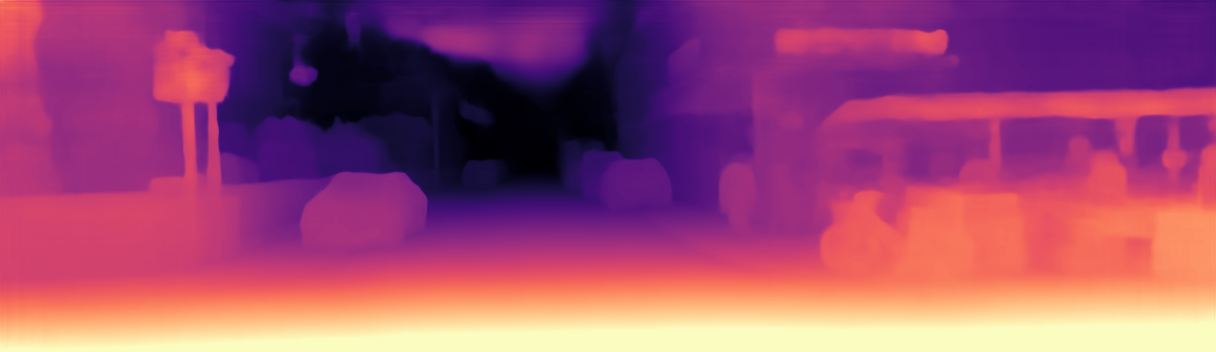}\\

% \rotatebox[origin=l]{90}{\makebox[0.22\columnwidth]{\scalebox{0.8}{DPT-Monodepth \cite{ranftl2021vision}}}}
% &\includegraphics[width = 0.8\columnwidth]{figures/compare/000dptmono.png}\\

\rotatebox[origin=l]{90}{\makebox[0.25\columnwidth]{\scalebox{1}{Monodepth2 \cite{godard2019digging}}}}
&\includegraphics[width = 0.95\columnwidth]{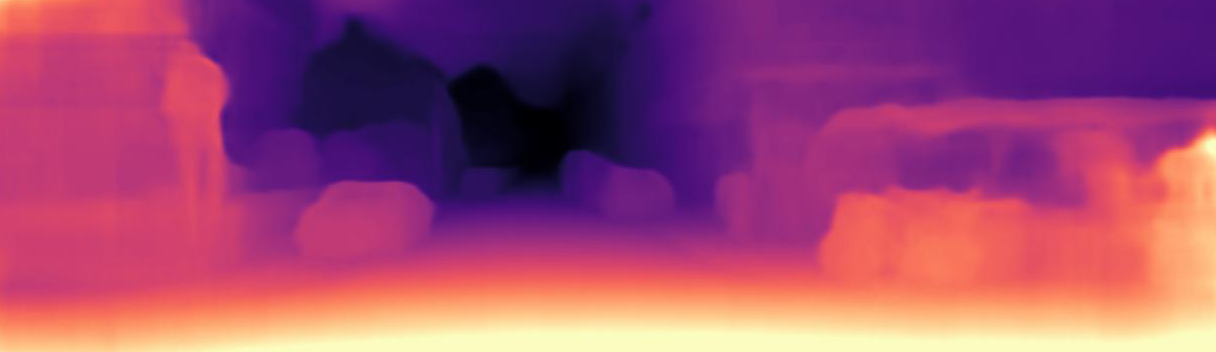}
\end{tabular}
\end{center}
\end{minipage}
\vspace{-2mm}
\caption{Depth estimation from cues. Monocular depth estimation relies on depth cues to make prediction. While our self-supervised stereo method, \textit{ChiTransformer}, leverages retrieved and rectified depth cues from a stereo to make superior predictions with context consistency.}
\vspace{-5mm}
\label{figure:compare}
\end{figure}

\vspace{-5mm}
\section{Introduction}
\label{sec:intro}
%Transformer has been proven to be very effective in various vision tasks. Recently, vision transformer is exploited to depth estimation and achieves superior results. In this paper, we push its capability further to stereo vision tasks and non-rectilinear settings. 

In the context of computer vision, nowadays almost all mainstream depth estimation methods are deep learning based and can be roughly categorized into two prevalent methodologies, namely, stereo matching and monocular depth estimation. Stereo matching has traditionally been the most investigated area due to its strong connection to the human visual system. The task is to find or estimate the correspondences of all the pixels in two rectified images ~\cite{barnard1982computational,scharstein2002taxonomy,andrew2001multiple}. Virtually, all the current works resort to convolutional neural network (CNN) based methods to calculate the matching cost since it was first introduced to the task by ~\cite{dosovitskiy2015flownet,zbontar2016stereo} in 2015. Following the seminal work of FlowNet~\cite{dosovitskiy2015flownet}, more than 150 papers have been published using CNN-related methods ~\cite{laga2020survey}, pushing the performance forward by more than 50\%. Some deep-seated issues such as thin structures, large texture-less areas, and occlusions have been mitigated or addressed over time~\cite{zhang2019ga, ilg2018occlusions}. So far, stereo matching is the most adopted technique in majority of passive stereo applications. 

However, the applications that entail depth estimation grow increasingly demanding as visual systems are greatly downsized and installed on platforms with higher mobility (e.g., UAV, commercial robots). This indicates a more congested, cluttered and dynamic operating environment where the once side issues become major ones, i.e., large disparity, severe occlusion and non-rectilinear images that might be involved. Therefore, most existing stereo matching methods are not set up for this new trend and fail to address these issues properly. 

On the other hand, monocular depth estimation (MDE) is spared from these challenges as depth is estimated from a single view. Following~\cite{eigen2014depth}, current works leverage deep models to derive more descriptive cues to achieve superior predictions. More recent works focus on fusing multi-scale information to further improve the pixel level depth estimation~\cite{liang2018learning,lee2019big}. Lately, vision transformer is exploited in the task and yields globally organized and coherent predictions with finer granularity~\cite{ranftl2021vision,bhat2021adabins}. State-of-the-art MDE methods can achieve impressive results with relative accuracy $\delta^3 > 0.99$ with supervised training~\cite{kittiMono1, kittiMono2, kittiMono3, kittimono4}. However, the reliability of MDE estimation is essentially based on the assumption that scenes in the real world are mostly regular. Therefore, due to the lack of stereo relation, the MDE is more delimited to its training dataset and susceptible to ``unfamiliar" scenes. This renders the MDE alone not reliable in safety-critical applications, such as autonomous driving and visual-aided UAV.

From the discussion above, we can see the limitations and advantages of stereo matching and MDE are complementary. Therefore, in this paper we propose a novel method that jointly addresses their limitations by crossing over the stereo and MDE approaches such that stereo information can be injected into the MDE process to rectify and improve the estimation made from depth cues.

We therefore introduce ChiTransformer, an optic-chiasm-inspired self-supervised binocular depth estimation network. ChiTransformer adopts the recent vision transformer (ViT)~\cite{vit} as backbone and extends the encoder-only transformer to an encoder-decoder structure similar to the ones for natural language processing~\cite{devlin2018bert, vaswani2017attention}. Rather than end-to-all connections, interleaved connections are employed for cross-attention to enable progressive instillation of the encoded depth cues from a nearby view to the master view in a self-regressive process. Our main contribution is the design of a retrieval cross-attention layer. Instead of attending over multi-level contextual relation of the encodings in the regular multi-head attention (MHA), the cross-attention mechanism of ChiTransformer aims to retrieve depth cues with strong contextual and feature coincidence from the other view. To achieve this, we condition the initial state (query) with a self-adjoint operator without breaking the convergence rule of modern Hopfield network~\cite{ramsauer2020hopfield}. The positive-definite operator is spectrally decomposed to enable polarized attention within the encoded feature space to emphasize on certain cues while preserving as much of the original information as possible. We show that this design facilitates reliable retrieval and leads to finer feature-consistent details on top of the globally coherent estimation. Moreover, the model can be further extended to non-rectilinear images such as fisheye by using a gated positional embedding~\cite{d2021convit}. We model the epipolar geometry with learnable quadratic polynomials of relative positions. Considering the per-pixel labeled data is challenging to acquire at scale and let alone for the non-rectilinear images, we choose to train the model with self-supervised learning strategy tailored from the work~\cite{godard2019digging}.

In contrast to traditional stereo methods, our approach foregoes pixel-level matching optimization but leverages the context-infused depth cues of both views to improve the overall depth prediction. With a global receptive field, ChiTransformer is not only not restricted to certain epipolar geometry, e.g., the horizontal collinear epipolar lines of rectified regular stereo pairs, but also able to treat large disparity. Furthermore, with the inherent capability of depth estimation within a single image, estimation at large occluded area can be properly handled rather than being masked out, interpolated, or left untreated. Enhanced from current MDE methods, our approach provides reliable prediction with guided cues in stereo pair which makes ChiTransformer more suitable for complex and dynamic environments.

Experiments are conducted on depth estimation tasks that provide stereo pairs. Our result shows that ChiTransformer delivers an improvement of more than 11\% compared to the top-performing self-supervised stereo methods. The architecture is also tested on stereo tasks to evaluate the gain brought by stereo cues and the underlying reliability yielded by the instilled stereo information. To show the potential of ChiTransformer in non-rectilinear images, we train our model to predict the distances on the translated synthetic fisheye sequences from~\cite{eichenseer2016data} and achieve visually satisfactory results.   

\begin{figure*}[t]
    \vspace{-5mm}
    \centering
    \includegraphics[width=1.0\textwidth]{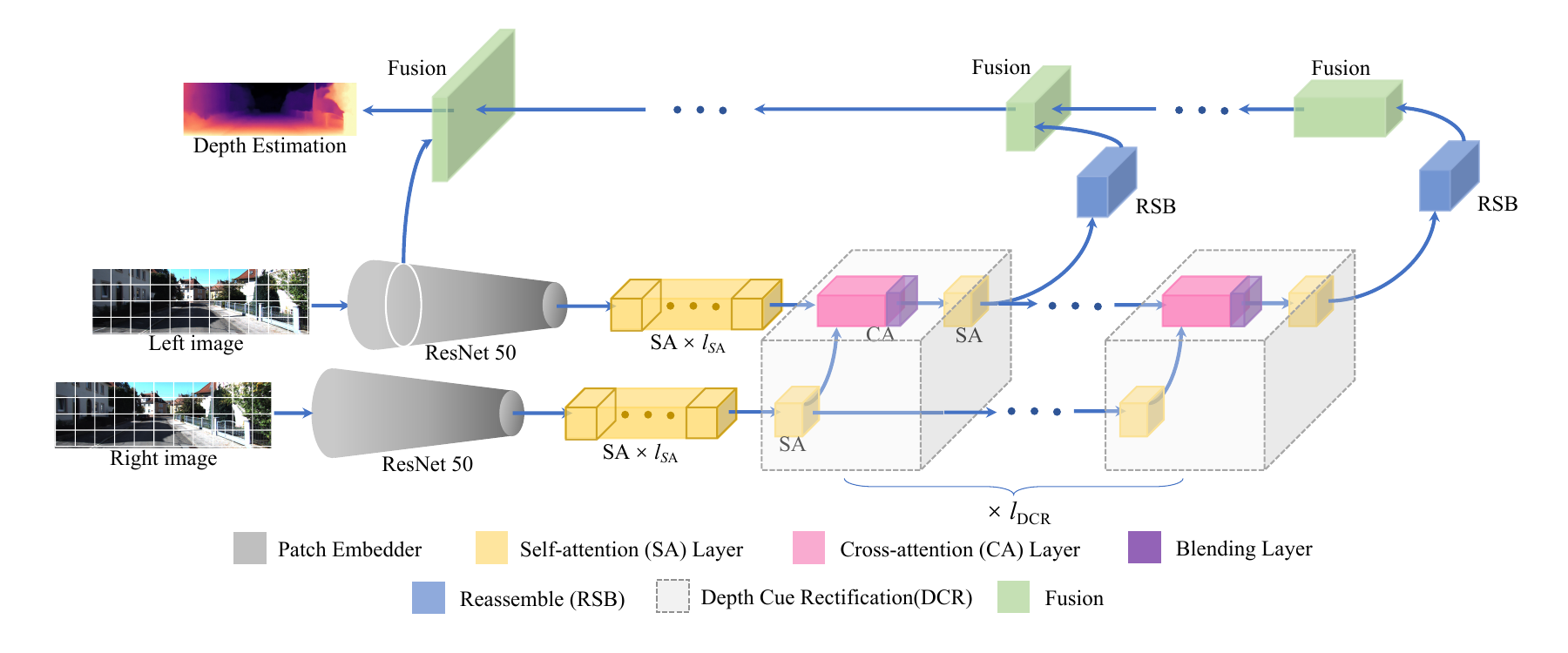}
    \vspace{-30px}
    \caption{\footnotesize Architecture of ChiTransformer. A stereo pair (left: master, right: reference) is initially embedded into tokens through a Siamese ResNet-50 tower. The 2D-organized tokens from the two images are flattened and then augmented with learnable positional embeddings and an extra class token, respectively. Then tokens are fed into two self-attention (SA) stacks of size $l_{\text{SA}}$ in parallel. After that, tokens are fed into a series ($\times l_{DCR}$) of depth rectification blocks (DCR) in each of which tokens of reference image go through an SA layer while tokens of master go through a polarized cross-attention (CA) layer, followed by an SA layer. In the polarized CA layer, relevant tokens from the output of reference SA are fetched to rectify the master's depth cues. Tokens from different stages are afterwards reassembled into an image-like arrangement at multiple resolution (blue) and progressively fused and up-sampled through fusion block to generate a fine-grained depth estimation.}
    \label{figure:arch}
    \vspace{-10px}
\end{figure*}

%\vspace{-3pt}
\section{Related Work}\label{sec: Rel.W}\vspace{-7pt}
Since the publications of~\cite{eigen2014depth,fischer2015flownet}, the end-to-end trainable CNN-based models have been the prototype architecture for dense depth~\cite{godard2017unsupervised,ranftl2019towards, godard2019digging} or disparity estimation~\cite{survey18, survey93, survey27, survey115}. The principal idea is to leverage learned representation to improve matching cost ~\cite{khan2018guide, scharstein2002taxonomy} or depth cues ~\cite{bhoi2019monocular} with contextual information of appropriately large local area. The prevalent encoder-decoder structure enables progressive down- and up-sampling of representation at different scales~\cite{diffscales1, chang2018pyramid,liang2018learning,yang2018segstereo,duggal2019deeppruner}, and the intermediate results from previous layers are often reused to recover fine-grained estimation while ensuring sufficiently large context. 

After showing exemplary performance on a broad range of NLP tasks, attention or in particular transformer has demonstrated competitive or superior capability in vision tasks such as image recognition~\cite{vit, touvron2021training}, object detection~\cite{carion2020end, zhu2020deformable}, semantic segmentation~\cite{ye2019cross}, super-resolution~\cite{yang2020learning}, image impainting~\cite{zeng2020learning}, image generation~\cite{razavi2019generating}, text-image synthesis~\cite{DaLLe}, etc. The successes also sparked interest in the community of stereo and depth estimation. ~\cite{wang2020parallax} leveraged cascaded attentions to calculate the matching cost along the epipolar lines and achieved competitive results among self-supervised stereo matching methods~\cite{zhou2017unsupervised,li2018occlusion, ahmadi2016unsupervised, jason2016back}. More recently, vision transformer was leveraged in place of convolution network as backbone for dense depth prediction in~\cite{ranftl2021vision} and achieved a significant improvement by 28\% compared to the state-of-the-art convolutional counterparts. A mini-ViT block~\cite{bhat2021adabins} is employed in the refinement stage to facilitate the adaptive depth bin calculation, and the work tops KITTI~\cite{kitti} and NYUv2~\cite{nyu} leaderboards. Inspired by~\cite{ranftl2021vision}, our method leverages the capability of ViT in learning long range complex context information to rectify depth cues instead of performing stereo matching.

Most works discussed above are fully supervised, which necessitate pixel-wise labeled ground truth for training. However, it is challenging to acquire dense annotations at scale in many real-world settings. One of the workarounds is to adopt self-supervised learning. For stereo self-supervised training, usually pixel disparities of synchronized stereo pairs are predicted~\cite{yang2018segstereo,zhou2017unsupervised,li2018occlusion, ahmadi2016unsupervised, jason2016back}, while for monocular self-supervised training, not only depth but also camera pose has to be estimated to help reconstruct the image and constrain the estimation network~\cite{byravan2017se3,zhou2017unsupervised,yin2018geonet,vijayanarasimhan2017sfm,godard2017unsupervised}. Considering the versatility and the potential application environment of our method, we choose self-supervised training for ChiTransformer. %Details are discussed in Section~\ref{sec:method}.

%-------------------------------------------------------------------------
\section{Method}
\label{sec:method}
This section introduces the overall architecture of the ChiTransformer with elaboration on the key building blocks. We follow the configuration of the vision transformer~\cite{vit} as the backbone and maintain the prevalent overall encoder-decoder structure because of their repeatedly verified success in various dense prediction tasks. We show the interplay of the encoded representations or cues between a stereo pair in ChiTransformer and how they can be effectively converted into dense depth prediction. The intuition for the elicitation and success of this method is discussed.

\subsection{Architecture}
\noindent\textit{\textbf{Overview:}} The complete architecture of ChiTransformer is shown in Figure~\ref{figure:arch}. ChiTransformer employs a pair of hybrid vision transformers as backbone with ResNet-50 ~\cite{he2016deep} for patch embedding. The parameters of two ResNet-50 are shared to ensure consistency in representations. The image patch embeddings are first projected to 768 dimensions, then flattened and summed with positional embeddings before fed into attention blocks. For an image of size $H\!\times\!W$, with the patch size of $P\!\times\!P$, the result is a set $T = \{t_0, t_1, \dots, t_{N_p}\}$, where $N_p = \frac{H\cdot W}{P^2}$ and $t_0$ is the class token. Here, patches are in the role of ``words" for transformer, but we will refer to patches as ``words" or ``tokens” interchangeably hereafter. The attention block for the reference view closely follows the design in~\cite{vit} with class token included, whereas master tokens are self-attended in the first multiple SA layers, followed by cross-attention (CA) and self-attention (SA) layers in an interleaved fashion. The output tokens of the master ViT (and reference ViT in training) are then reassembled into an image-like arrangement. Feature representations $I^s$ at different scales $s \in S$ are progressively aggregated and fused into the final depth estimation in the fusion block, which is modified from RefineNet~\cite{lin2017refinenet}. Fusion block is shared for both views in training phase but dedicated to the master view in inference. \\
%with CA layer being inserted after every other SA layer

\noindent\textit{\textbf{Attention layers:}} Self-attention layer is the crucial part for transformers and other attention-based methods to achieve superior performance over their non-attention competitors. The key advantage is that complex context information can be gathered in a global scope. With multiple layers of SA, encodings get progressively tempered with the context information as it goes deeper into the attention layers. This mechanism begets the globally coherent predictions. Therefore, instead of putting immediate connection to the CA layer, we place multiple ($l_{SA} =  4$) SA layers at the output of the ResNet-50. Cues with appropriate amount of context information result in more reliable pattern retrieval in the subsequent CA layers. This design improves both training convergence and prediction performance.

The \textit{cross-attention} layer is our key contribution in ChiTransformer. It is the enabler of the stereopsis through the fusion of high-level depth cue expressions from two views. We argue that the effectiveness of the traditional 4-step strategy would be largely weakened as the sources of ill-posedness, such as occlusion, wider and closer range of depth, depth discontinuity and nonlinearity, become increasingly frequent or prominent. Current deep learning-based methods rely on the learned rich representations to construct cost volume which is then regularized to make estimation. The output quality, in this case, largely depends on both the quality of the representations and the conformity of the scene to the matching regularizing assumptions~\cite{scharstein2002taxonomy}. While good representations can be learned with many approaches, there are few ways to fix up an impaired cost volume when scenes are far away from being appropriate for stereo matching. Therefore, instead of clinging to the matching strategy, we propose a novel pattern retrieval mechanism inspired by associative memory to retrieve the correspondent pattern from the other view. We assume that a set of patterns can be learned to separate well such that each pattern can be retrieved at least in meta-stable state, i.e., fixed average of similar patterns~\cite{ramsauer2020hopfield}. Modeled by modern Hopfield network~\cite{demircigil2017model,krotov2016dense}, the retrieval rule of the associative memory elegantly coincides with the attention mechanism of the transformer. Naturally, we leverage cross-attention layer to retrieve patterns (tokens) from the reference to the master view. To facilitate reliable effective retrieval, we devise a new attention mechanism -- polarized attention, which enables feature-sensitive retrieval while preserving the context information contained in the pattern without breaching the convergence rule. From~\cite{wang2020parallax}, we observe that direct attention over representations at the output of CNN reduces to cosine similarity-based matching. Without position-dependent context information over extensive scope, patterns are liable to ill-posedness and low separability.

Given a token pair $(^mt_i,\ ^mt^{\prime}_i)$, $\forall i$ $\in$ $\{1,$ $\cdots,$ $ N_p\}$, from the preceding CA layer and the class token pair $(^mt_0,\ ^rt_0)$, where $m$ indicates the master view, $r$ the reference view, and $t'$ denotes the retrieved tokens, depth cues are then rectified through the following blending process:
\begin{align}
    &\mathit{f}_{proj}(^mt_i,\ ^mt^{\prime}_i) = \text{MLP}(\left[^mt^{\top}_i,
    ^mt^{\prime\top}_i\right]),
\end{align}
\begin{align}
    &^mt_i =\ ^mt_i + \texttt{Heat}(p_{a_i})\cdot \mathit{f}_{proj}(^mt_i,\ ^mt^{\prime}_i).
\end{align}
We set $^mt^{\prime}_0 = {^r}t_0$ to unify the expression. GELU~\cite{hendrycks2016gaussian} is used for MLP nonlinearity, $p_{a_i}$, the vector of attention scores of $^mt_i$, and $\texttt{Heat}$, the confidence score calculated with stabilized attention entropy as
\begin{equation}
   \texttt{Heat}(p_{a_i}) = 1 - g\left(\text{H}(p_{a_i}), \tau, c\right),
\end{equation}
where $\text{H}(p_{a_i}) = -\sum^{N_p}_{k=1}{p_{a_{i, k}}\log{(p_{a_{i, k}} + \epsilon)}}$, and $g(\cdot)$ is a clamping function (e.g., sigmoid or smoothstep) with temperature $\tau$ and offset $c$. \texttt{Heat} is set to 1 for class token. By doing so, tokens retrieved back in fixed state, i.e., with a very low entropy, would be securely rectified whereas those with high entropy are inhibited from being updated as they are very likely to reside in occluded areas. Thus, the depth in occluded or ``uncertain” areas are left to the power of SA layers to speculate its value with context information and rectified cues from neighboring non-occluded areas. 

\vspace{3mm}
\noindent \textit{\textbf{Fusion block:}}
Our convolutional decoder follows the refinement block in~\cite{lin2017refinenet,ranftl2021vision}. The output of attention layers $\boldsymbol{t} \in {\rm I\!R}^{(N_p + 1)\times D}$ is reassembled into an image-like arrangement ${\rm I\!R}^{H^\prime \times W^\prime \times D'}$ through a four-step operation:
\begin{equation}
    \text{RSB} = \left(\text{rescale}\circ\text{reshape}\circ\text{MLP} \circ \text{cat} \right).
\end{equation}
The class token is concatenated with all other tokens (by broadcasting) before being projected to dimension $D^\prime$ to get $\bs{t}_{\setminus 0}$. Then it is reshaped into a 2D shape per the original arrangement of the image embedding. Finally, $\boldsymbol{t}_{\setminus 0}$ is re-sampled to size $\frac{H}{P}s_l \times \frac{W}{P}s_l \times D_l $ for different scales at level $l$. Re-sampling method is 2D transposed convolution for $s_l>1$ (up-sampling), and strided 2D convolution for $s_l<1$ (down-sampling). For our model, features from level $l_{attn} = \{11, 7, 3\}$ in attention blocks (12 in total for ChiTransformer-8) and level $l_{res} = \{1, 0\}$ (first 2 blocks) in ResNet-50 are reassembled. The reassembled feature maps from those levels are consecutively fused through customized feature fusion block from RefineNet~\cite{lin2017refinenet}. At each level, feature map is up-sampled by a factor of 2 and finally the depth estimation map reaches the original resolution of the input images.

The architecture of ChiTransformer is structurally similar and biologically analogous to the \textit{optic-chiasma} structure in our visual system, where visual field covered by both eyes is fused to enable the processing of binocular depth perception by stereopsis~\cite{scharstein2002taxonomy}, hence the name of our model.

\subsection{Polarized Attention}
We propose a new attention mechanism to highlight or suppress features, which is much like signal polarization but in feature domain. Ideally, for a set of tokens represented in tensor  $\bs{t}=(t_1, \cdots, t_N)$ that is well separated, highlighting or suppressing can be potentially achieved in token-wise granularity. However, ideal separability is hard to achieve in practice because the attention tensor $\bs{A}$ for regular attention mechanism is calculated as
\vspace{-1mm}
\begin{equation}
	\boldsymbol{A} = \text{softmax}\left(\beta\boldsymbol{t}^\top   \mathbf{W_t}^\top \mathbf{W}_{\xi}\boldsymbol{\xi}\right),
\vspace{-1mm}
\end{equation}	
which is prone to be noisy with joint activation over all channels and hard to learn directly for $\mathbf{W}$’s. While the prevalent MHA seeks for multi-level context instead of retrieval since tokens are mapped to different (sub-)spaces for each head that generates its own attention weights and output with the projected tokens. To achieve retrieval behavior, without loss of generality, we stick to the Hopfield network update rule to ensure retrieval behavior and condition the query pattern with a self-adjoint operator $\mathbf{G} \in \rm{I\!R}^{\mathit{d\times d}}$,
\vspace{-2mm}
\begin{equation}
\boldsymbol{\xi}^\prime = \boldsymbol{t}\text{softmax}\left(\beta\boldsymbol{t}^\top   \mathbf{G} \boldsymbol{\xi} \right),
\vspace{-2mm}
\end{equation}
where $\beta$ is the scale factor set to be $1/\sqrt{d}$. We assume the constraint that the query and memory should stay in the same sub-space which is satisfied by a positive-definite $\mathbf{G}$ decomposed as $\mathbf{G} = \mathbf{M}^\top \mathbf{M}$. It can further be spectrally decomposed to get: 
\begin{equation}
\boldsymbol{\xi}^\prime = \boldsymbol{t}\text{softmax}\left(\beta\boldsymbol{t}^\top   \mathbf{U}^\top\mathbf{\Lambda} \mathbf{U}\boldsymbol{\xi} \right),
\vspace{-2mm}
\end{equation}
where $\mathbf{U}$ is an orthogonal matrix and $\mathbf{\Lambda}$ is a positive diagonal matrix.

To achieve feature-sensitive retrieval while factoring in all the information in the embeddings, we desire $\text{diag}(\mathbf{\Lambda})$ not to be zero abounded, i.e., feature selection. To achieve that and also enable multi-modal retrieval, multiple $\mathbf{\Lambda}$s are learned and we desire $\prod^{s}_{i=1}{\mathbf{\Lambda}_i}$ to be close to $\mathbf{I}$ such that if one feature is highlighted in one mode it should be suppressed in other modes. As such, the new attention mechanism becomes
\begin{equation}
    \boldsymbol{\xi}^\prime = \mathbf{W}\overset{s}{\underset{i=1}{\texttt{cat}}}\left[\boldsymbol{t}\cdot\text{softmax}\left(\beta\boldsymbol{t}^\top   \mathbf{U}^\top\mathbf{\Lambda}_i\mathbf{U}\boldsymbol{\xi} \right)\right].
\end{equation}
% \begin{equation}
% \begin{split}
%     \boldsymbol{\xi}^\prime = \sum^{s}_{i=1}\mathbf{M}^R_i\odot\left(\boldsymbol{t}\cdot\text{softmax}\left(\beta\boldsymbol{t}^\top   \mathbf{U}^\top\mathbf{\Lambda}_i\mathbf{U}\boldsymbol{\xi} \right)\right),\\
%     \mathbf{M}_i^R = 
%     \begin{cases}
%       1 & i = j,\ j = \text{argmax}_i\ \texttt{Heat}\left(p_{a_{i}}\right)\\
%       0 & i\neq j
%     \end{cases}      
% \end{split}
% \end{equation}

\begin{figure}[h]
%\vspace{-5mm}
\begin{minipage}{\columnwidth}
    \setlength{\tabcolsep}{0pt}
    \renewcommand{\arraystretch}{0.1}
\begin{center}
\scriptsize
\begin{tabular}{c@{}c@{}}

%\rotatebox[origin=l]{90}{\scalebox{0.7}[0.7]{Master View}} \hspace{1pt}
\includegraphics[width = 0.5\columnwidth]{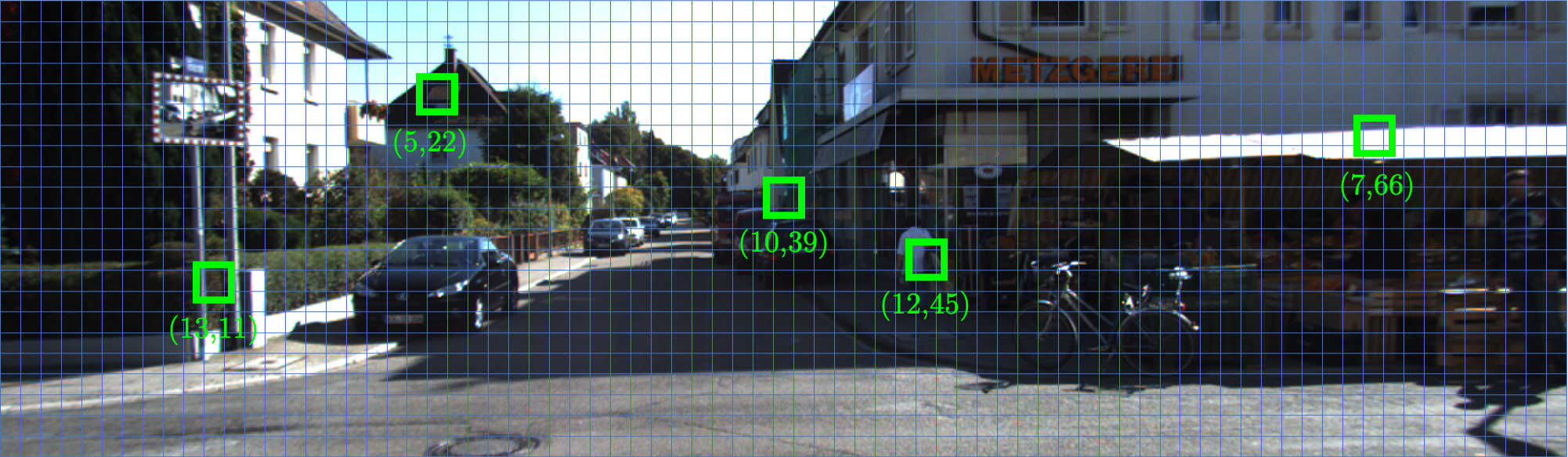} 
& \includegraphics[width = 0.5\columnwidth]{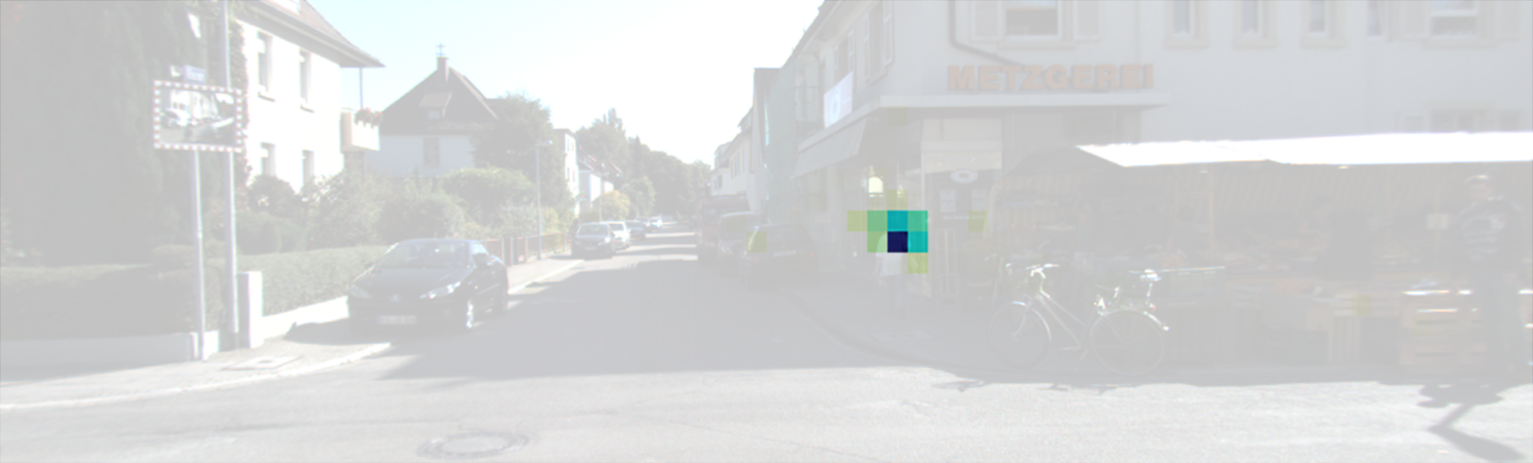}\\[-3mm]

\makebox[1in]{\textcolor{yellow}{\tiny{Reference (right image)}}}
& \makebox[1in]{(12,45)}\\[2mm]

\includegraphics[width = 0.5\columnwidth]{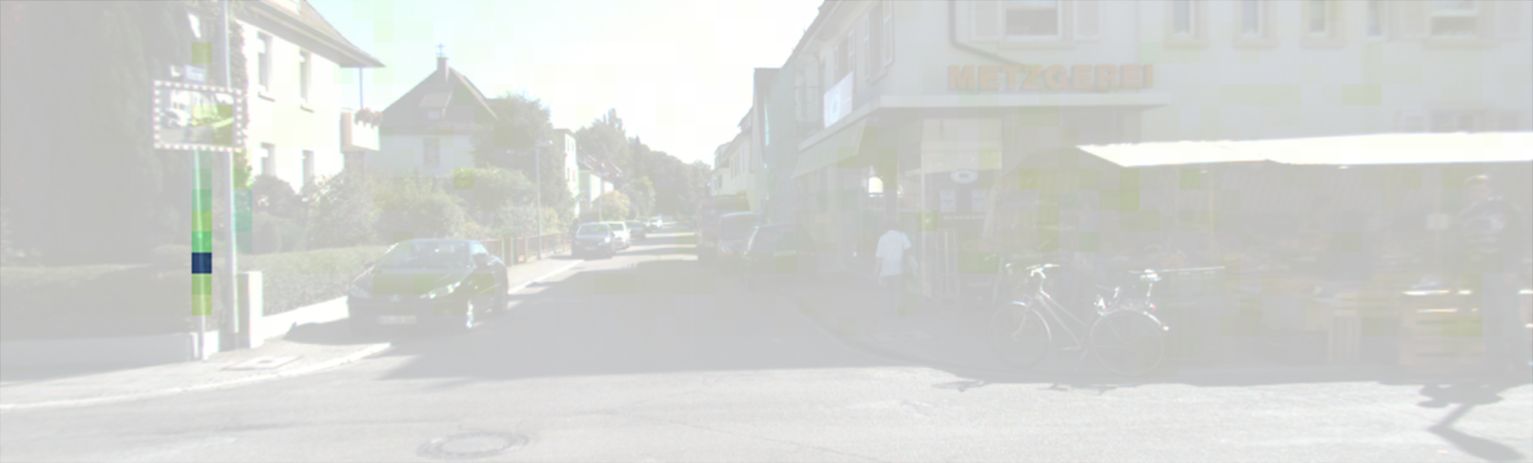} 
&\includegraphics[width = 0.5\columnwidth]{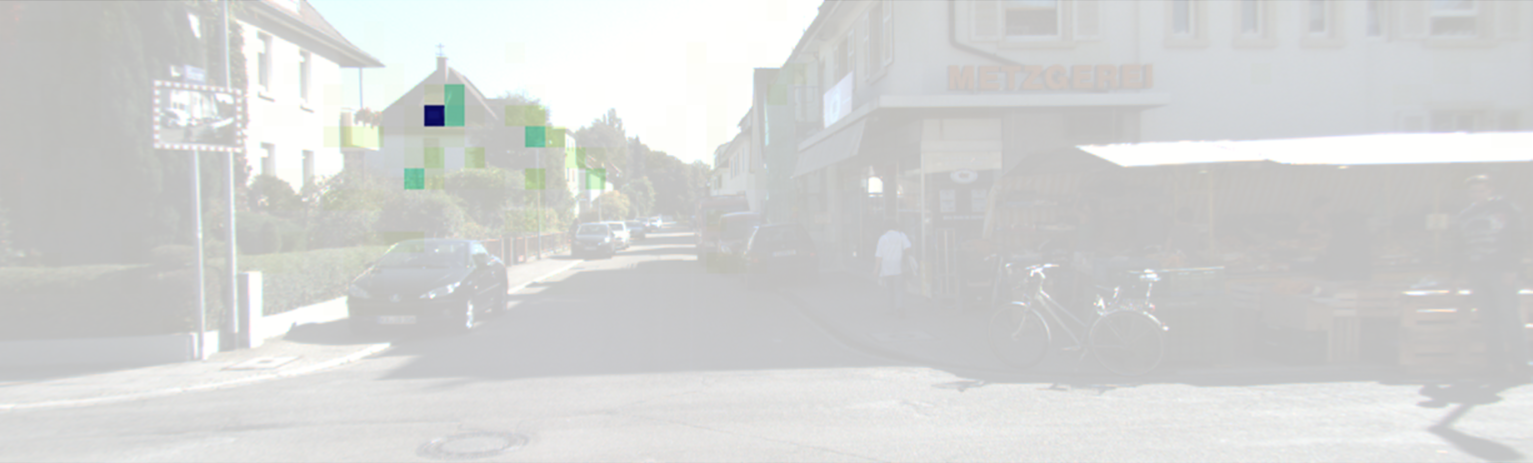} \\[-3mm]

\makebox[1in]{(13,11)}
& \makebox[1in]{(5,22)}\\[2mm]

\includegraphics[width = 0.5\columnwidth]{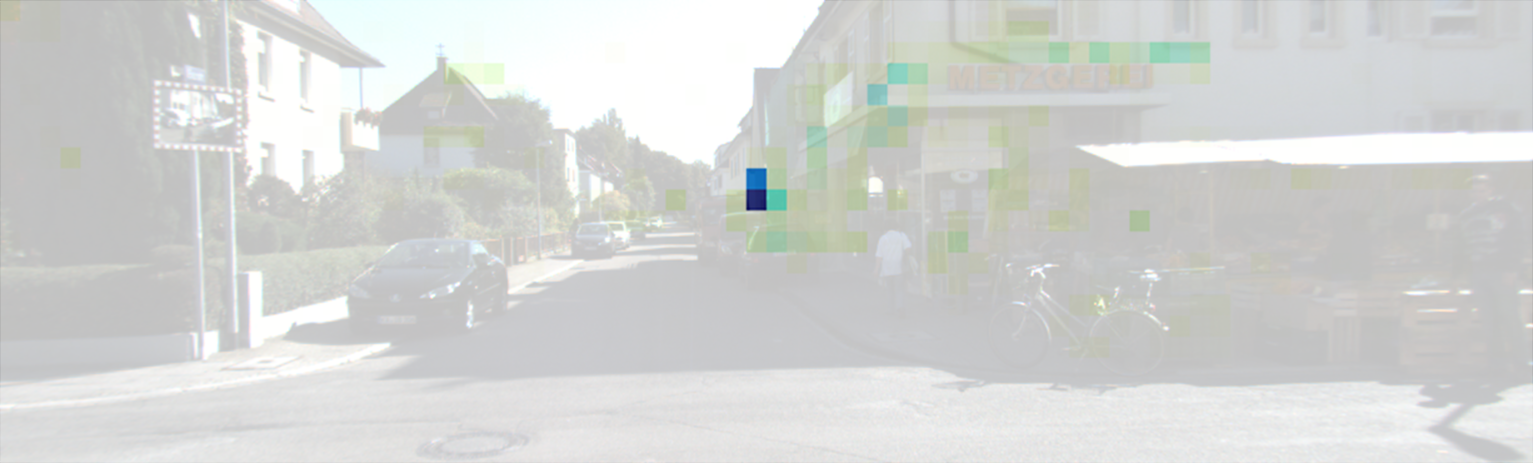} 
&\includegraphics[width = 0.5\columnwidth]{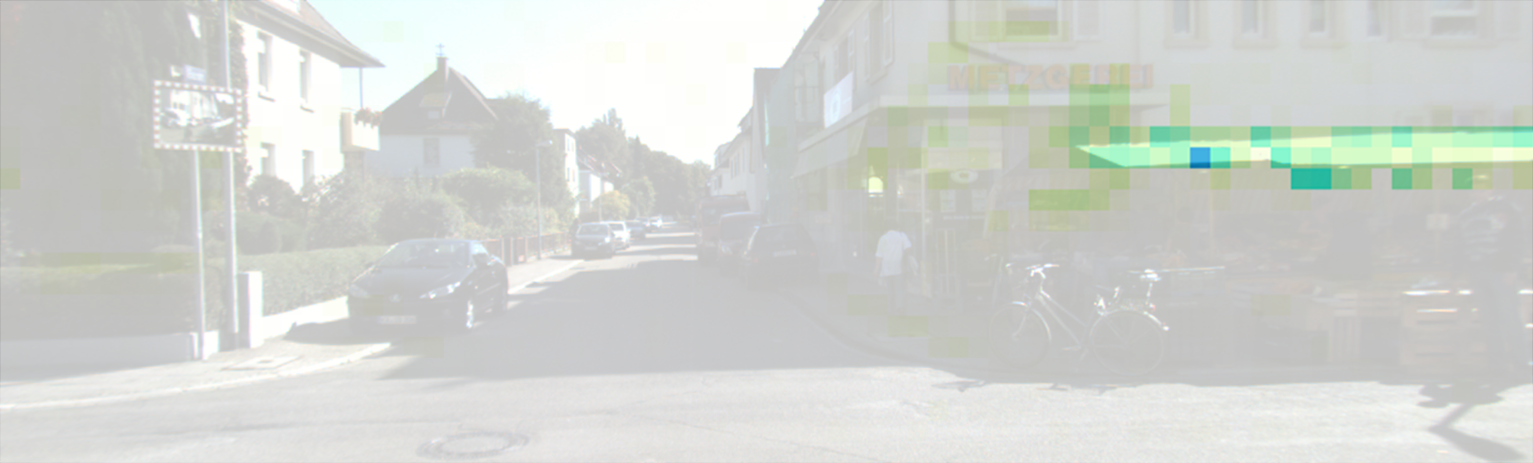} \\[-3mm]

\makebox[1in]{(10,39)}
& \makebox[1in]{(7,66)}\\[2mm]

\end{tabular}
\end{center}
\end{minipage}
\vspace{-2mm}
\caption{Attention map of $\boldsymbol{A}^{s=1}_{ij}$ of tokens retrieved in meta-stable state in the CA layer of the $2^{nd}$ DCR. Darker green corresponds to higher attention score. As expected, tokens of distinctive features have more centralized attention, while tokens in texture-less area result in more dispersed attention.}
\vspace{-3mm}
\label{figure:attn_compare}
\end{figure}

For our model, $\bs{\xi}$ is the tokens from the master view, $\bs{t}$ is the tokens from reference view, $\mathbf{W}$ projects the concatenated tokens back to its original dimension, and $\bs{\xi}^\prime$ is the retrieved tokens from the reference view. The lowest entropy of the $s$ heads is used in \texttt{Heat} calculation for rectification. The effectiveness of the retrieval mechanism is reflected in Figure~\ref{figure:attn_compare}, where up-scaled attention map of different tokens are overlapped with the reference view.

%%%%%%%%%%%Attention Map%%%%%%%%%%%%
%%%%%%%%%%%%%%%%%%%%%%%%%%%%%%%%%%%%

\subsection{Learnable Epipolar Geometry}
Token separability may be limited by the memory size and image content (e.g., existence of repetitive or uniform texture in the image). To further ensure secure retrieval without corrupting the encoded information, we constrain the attention mechanism with epipolar geometry through a \textit{gated positional cross attention} (GPCA) following~\cite{d2021convit}. In GPCA, positional embedding is modeled as trainable quadratic polynomial of relative positional encoding $\bs{v}_{pos}^{\top}\bs{r}_{ij}$~\cite{cordonnier2019relationship}. For regular rectified stereo, candidate retrievals reside within collinear horizontcal lines. Therefore, we set $\bs{v}_{pos}  = -\alpha\left(0, 0, 0, 0, 0, 1, \cdots, 0\right)$, $ 
\bs{r} = \left(1, \delta_1, \delta_2, \delta_1\delta_2, \delta_1^2, \delta_2^2, 0, \cdots, 0\right)$. For non-rectilinear images, e.g., fisheye, $\bs{v}_{pos}$ is a vector of trainable curve coefficients, which will be discussed when we present results on fisheye images.

In the equations above, $\boldsymbol{r}$ is the position vector of $(\delta_1, \delta_2)$ which are the relative coordinates with respect to the query. The locality strength $\alpha > 0$ determines how focused attention is along the horizontal line (i.e., when $\delta_2 = 0$). 
The positional attention scores are calculated as softmax normalized $L_2$ distance between the attended tokens and the query:
\begin{equation}
    \bs{A}_{pos,ij} = \text{softmax}\left(\bs{v}_{pos}^{\top}\bs{r}_{ij}\right).
\end{equation}
With the learnable gating parameter $\lambda$, the GPCA attention scores are calculated as:
\begin{equation}
    \boldsymbol{A}^s_{ij} = \texttt{norm}\left[(1-\sigma (\lambda))\boldsymbol{A}^s_{cnt,ij} + \sigma(\lambda)\boldsymbol{A}_{pos,ij}\right],
\end{equation}
where $\texttt{norm}[\mathbf{x}] = \frac{x_{ij}}{\sum_{k}{x_{ik}}}$,  $\sigma$ is the sigmoid function, and $\bs{A}^s_{cnt,ij}$ is the content attention score calculated by polarized attention of head $s$.
To avoid GPCA from being stuck at $\lambda >> 1$, we initialize $\lambda=1$ for all layers.

\subsection{Regularization}
Matrix $\mathbf{U}$ has to be orthogonal to guarantee that query and memory are attended in the same space. However, $\mathbf{U}$ in each layer is trainable parameter; even though it can be initialized with orthogonal matrices, during training process the orthogonality may not hold. Therefore, we introduce an orthogonality regularization loss to $\mathbf{U}$ as:
\vspace{-2mm}
\begin{equation}
    L_o(\mathbf{U}) = \frac{1}{d^2}\left\|\mathbf{U}^{\top}\mathbf{U} -  \mathbf{I}\right\|_{\mathcal{F}},
\vspace{-2mm}
\end{equation}
where $d$ is the size of $\mathbf{U}$ and $\left\|\cdot\right\|_{\mathcal{F}}$ is the Frobenius norm of matrix. Although $\mathbf{U}$ can be orthogonalized through Cayley's parameterization, it is computational expensive for large matrix as inversion is involved and we found it is more difficult to converge and unstable in our case.

To induce the diagonal matrix $\mathbf{\Lambda}$ to be trained into the desired form, we modified Hoyer regularizer~\cite{hoyer2004non} to mitigate the proportional scaling issue and at the same time to pull $\mathbf{\Lambda}$ away from being identity matrix. We introduce the following regularization:
\vspace{-1mm}
\begin{equation}
    L_\Lambda(\mathbf{\Lambda}) = \frac{\left|\prod^{s}_{i=1}{\left|\mathbf{\Lambda}_i\right|_e} - \mathbf{I}\right|_1}{\prod_{i=1}^{s}{\left\|\mathbf{\Lambda}_i\right\|_{\mathcal{F}}}},
\vspace{-1mm}
\end{equation}
where $|\cdot|_e$ is the element-wise absolute function. Identity matrix is only one of the possible but unpreferred solution for $\prod^{s}_{i=1}{|\mathbf{\Lambda_i}|_e} - \textbf{I} = 0$. Take 2D matrix for example, diagonal matrices $\Lambda_1 = (5, \frac{1}{5})$, $\Lambda_2=(\frac{1}{5},5)$ is also a solution. However, when $\Lambda_i$s are approaching to this optimal, the denominator of Eq.~12 is much larger ($\sim\!\!25$) than that of $\Lambda_i$s close to the identity matrix (2.0). In addition, the final loss is a combination of reconstruction loss and regularization loss; with a proper hyperparameter $\mu_\lambda$, the learned $\mathbf{\Lambda}$s will be pulled away from \textbf{I}.

\subsection{Training}
In this section, we provide details of the training method we used. We closely followed the self-supervised stereo training method provided in~\cite{godard2019digging}. The model is trained to predict the target image from the other viewpoint in a stereo pair. Unlike classical binocular and multi-view stereo methods, the image synthesis process in our case is constrained by predicted depth instead of disparity as an intermediary variable. Specifically, given a target image $I_t$, a source image $I_t'$, and the predicted depths $D_t$, through the relative pose between two views $T_{t\rightarrow t^\prime}$ calculated with the provided stereo base width ($0.54m$ for KITTI) and calibration information, the correspondent coordinates between two images can be calculated. Following~\cite{jaderberg2015spatial}, the target image can be reconstructed from source image using bilinear sampling, which is sub-differentiable.

The depth prediction should minimize the photometric reprojection error constructed for both master and reference view as follows:
\begin{align}
    \vspace{-1mm}
    L_p = \omega\cdot\texttt{pe}\left(I_t, I_{t'\rightarrow t}\right) + (1-\omega)\cdot\texttt{pe}\left(I_{t'}, I_{t\rightarrow t'}\right),
    \vspace{-1mm}
\end{align}
where $\omega$ is the weight for master view, and $\texttt{pe}(\cdot)$ is the photometric reconstruction error~\cite{wang2004image}:
\begin{align}
    \texttt{pe}(X, Y) = &\frac{\kappa}{2}\left(1-\text{SSIM}\left(X, Y\right)\right) \nonumber\\ &+ (1 - \kappa)\left\|X - Y\right\|_1
\end{align}
$\kappa$ = 0.85 and $I_{t^\prime\rightarrow t}$ is the reprojected image:
\begin{align}
    I_{t^\prime\rightarrow t} = \texttt{bi-sample}\left<\texttt{proj}\left(D_t, T_{t\rightarrow t^\prime}, K\right)\right>,
\end{align}
where $K$ is the pre-computed intrinsic matrix, \texttt{proj} is the resulting image coordinates projected from source view through \vspace{-10pt}
\begin{align}
    p_t^\prime := KT_{t\rightarrow t^\prime}D_t[p_t]K^{-1}p_t
\end{align}
and $\texttt{bi-sample}\left<\cdot\right>$ is the bilinear sampler.

We also enforce edge-aware smoothness in the depths to improve depth-feature consistency defined as
\begin{align}
    L_s = |\partial_xd_t^\ast|e^{-|\partial_xI_t|} + |\partial_yd_t^\ast|e^{-|\partial_yI_t|},
\end{align}
where $d_t^\ast\!\! =\!\! d/\Bar{d}_t$ is the mean-normalized inverse depth in~\cite{wang2018learning}. 

Unlike existing self-supervised stereo-matching methods that rely on predicted values to generate confidence map to detect occlusions, e.g. left-right consistency check, ChiTransformer detects occluded area on the fly in the form of heat map in the rectification stage. During training, heat map from the last GPCA layer is up-sampled to the output resolution and used as a mask $m_h$ in loss computation. For stereo training, static camera and synchronous movement between objects and camera are not issues, hence we do not apply the binary auto-masking to block out the static area in the image.

During inference, only the master ViT output is up-scaled and refined to make the prediction. While in the training stage, both ViT towers in ChiTransformer are trained in tandem to predict depth and calculate losses $L_p$ and $L_s$.

\paragraph{Final Training Loss}

By combining the reconstruction loss, per-pixel smoothness from two views and the regularizations for the matrices $\mathbf{U}$ and $\mathbf{\Lambda}$, the final training loss is:\vspace{-5pt}
\begin{equation}
    \mathcal{L} = \texttt{mean}(m_h\odot L_p) + \mu_s L_s + \mu_o L_o + \mu_\lambda L_\lambda,
\end{equation}
where $\mu_*$ are the hyperparameters that balance the contributions from different loss terms.

Our models are implemented in PyTorch. With pretrained ResNet-50 patch feature extractor and partial refinement layers from~\cite{ranftl2021vision}, the model is trained for 30 epochs using using Adam~\cite{kingma2014adam} with a batch size of 12 and input resolution of $1216 \times 352$. We use learning rate $1e\!-\!5$ for the ResNet-50 and $1e\!-\!4$ for the rest part of the network in the first 20 epoch, and then is decayed to $1e\!-\!5$ for the remaining epochs. We set $\omega\!=\!0.6$, $\mu_s=1e\!-\!4$, $\mu_o=1e\!-\!7$ and $\mu_{\lambda}=1e\!-\!3$ in our experiments.  

%-------------------------------------------------------------------------
\vspace{-10pt}
\begin{table}[b]
\footnotesize
\caption{\vspace{-3mm}\textsc{Quantitative Results}}
\label{table:stereo}
\begin{tabularx}{\columnwidth}{p{1mm}|l p{5mm}p{5mm}p{5mm} p{5mm}p{5mm}p{5mm}}
\thickhline
 & \multicolumn{1}{c}{Method}&\multicolumn{3}{c}{NOC}&\multicolumn{3}{c}{ALL}\\
 &            & D1 (bg)    &D1 (fg)    &D1 (all)   & D1 (bg)    &D1 (fg)    &D1 (all)\\
\hline
\parbox[t]{1pt}{\multirow{4}{*}{\rotatebox[origin=c]{90}{Supervised}}}
 & DispNet~\cite{mayer2016large}  & 4.11     & 3.72    & 4.05    & 4.32   & 4.41    & 4.43\\
 & GC-Net~\cite{kendall2017end}   & 2.02     & 5.58    &2.61     &2.21    & 6.16    & 2.87\\
 & iResNet~\cite{liang2018learning}  & 2.07     & 2.76    &  2.19   &  2.25  & 3.40    & 2.44\\
 & PSMNet~\cite{chang2018pyramid}   & 1.71     & 4.31    & 2.14    & 1.86   & 4.62    & 2.32\\
\hline
\parbox[t]{3mm}{\multirow{10}{*}{\rotatebox[origin=c]{90}{Self-supervised}}}
 &Yu et al.~\cite{jason2016back}   & -     & -      & 8.35   & -     & -     & 19.14\\
 &Zhou et al.~\cite{zhou2017unsupervised} & -     & -      & 8.61   & -     & -     & 9.91\\
 &SegStereo~\cite{yang2018segstereo}   & -     & -      & 7.70   & -     & -     & 8.79\\
 &OASM~\cite{li2018occlusion}        & 5.44  & 17.30  & 7.39   & 6.89  & 19.42 & 8.98\\
 &PASMnet_192~\cite{wang2020parallax} & 5.02  & 15.16  & 6.69   & 5.41  & 16.36 & 7.23\\
 &Flow2Stere~\cite{liu2020flow2stereo} & 4.77  & 14.03  & 6.29   & 5.01   & 14.62 & 6.61\\
 &pSGM~\cite{lee2017memory}        & 4.20  & 10.08  & 5.17   & 4.84   & 11.64  & 5.97\\
 &MC-CNN-WS~\cite{tulyakov2017weakly}   & 3.06  & 9.42   & 4.11   & 3.78   & 10.93  & 4.97\\
 &SsSMnet~\cite{diffscales1}     & 2.46  & 6.13   & 3.06   & 2.70   & 6.92   & 3.40\\
 &PVSstereo~\cite{wang2021pvstereo}   & \textbf{2.09}  & 5.73   & 2.69   & \textbf{2.29}  & 6.50  & 2.99\\
 
 &\textbf{ChiT-8} (ours)       & 2.24  & 4.33   & 2.56   & 2.50    & 5.49   & 3.03\\
 &\textbf{ChiT-12} (ours)   & 2.11  & \textbf{3.79}   & \textbf{2.38}  &2.34    & \textbf{4.05}    & \textbf{2.60}\\
\thickhline
\end{tabularx}
\begin{minipage}{\linewidth}\footnotesize
    \vspace{1mm}
     Comparison of our model to the state-of-the-art self-supervised binocular stereo methods. The lower the better for all metrics. 
\end{minipage}
\vspace{-5mm}
\end{table}

\renewcommand{\arraystretch}{1}
%-------------------------------------------------------------------------

\begin{figure*}[t]
\hspace{2mm}
\begin{minipage}{0.9\textwidth}
    \renewcommand{\arraystretch}{0.1}
    \begin{center}
    \scriptsize
    \begin{tabular}{@{}c@{}c@{}c@{}c@{}}
    \rotatebox[origin=l]{90}{\scalebox{0.7}[0.7]{\hspace{5mm}Left Image}} \hspace{1pt}
    & \includegraphics[width = 0.35\textwidth]{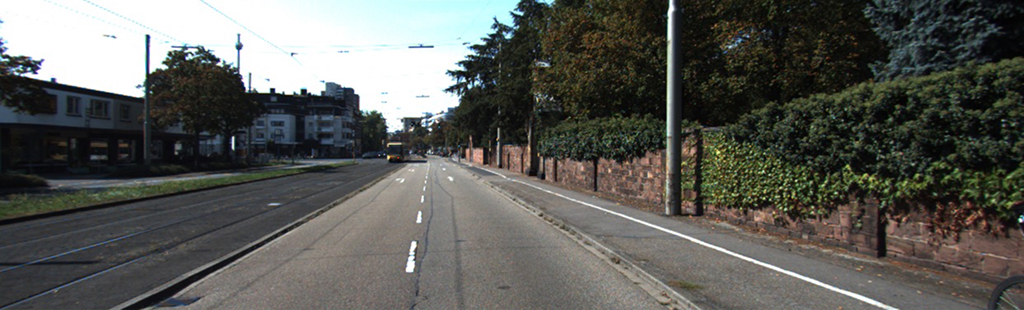}
    & \includegraphics[width = 0.35\textwidth]{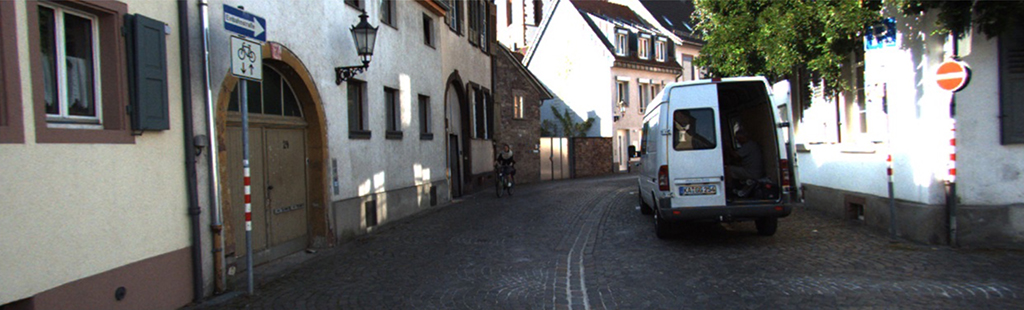} 
    & \includegraphics[width = 0.35\textwidth]{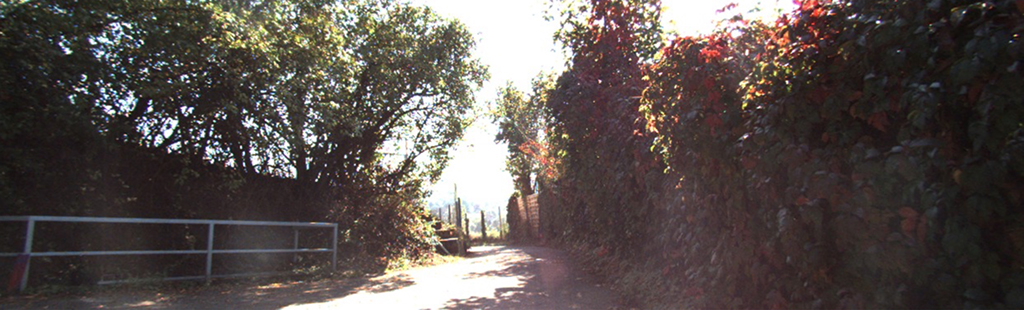}\\
    
    \rotatebox[origin=l]{90}{\centering\scalebox{0.7}[0.7]{\hspace{5mm}Monodepth2}} \hspace{1pt}
    & \includegraphics[width = 0.35\textwidth]{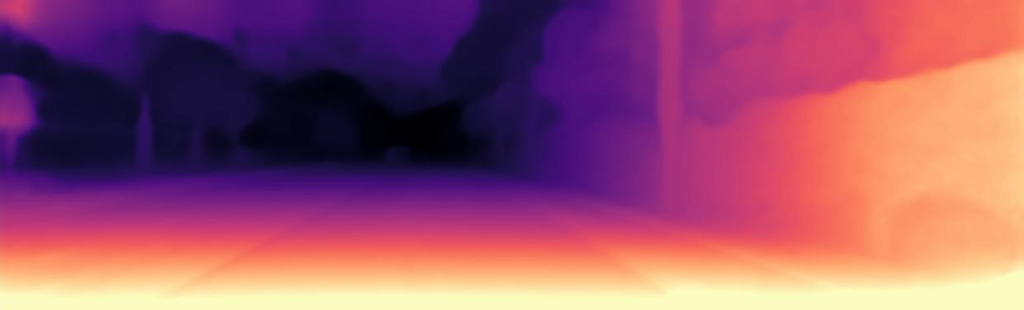} &
    \includegraphics[width = 0.35\textwidth]{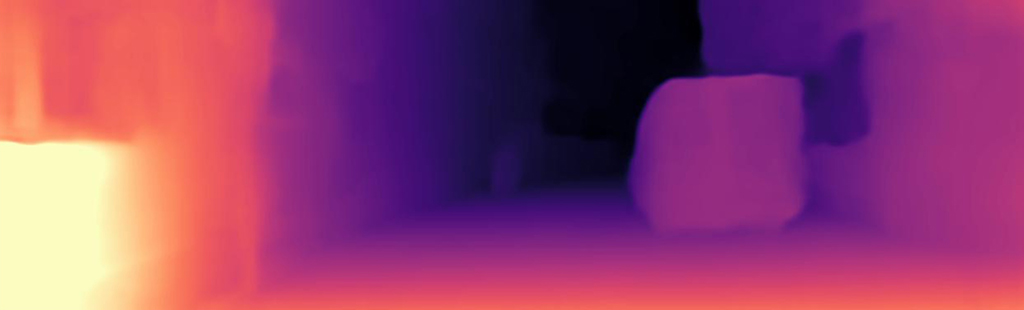} & \includegraphics[width = 0.35\textwidth]{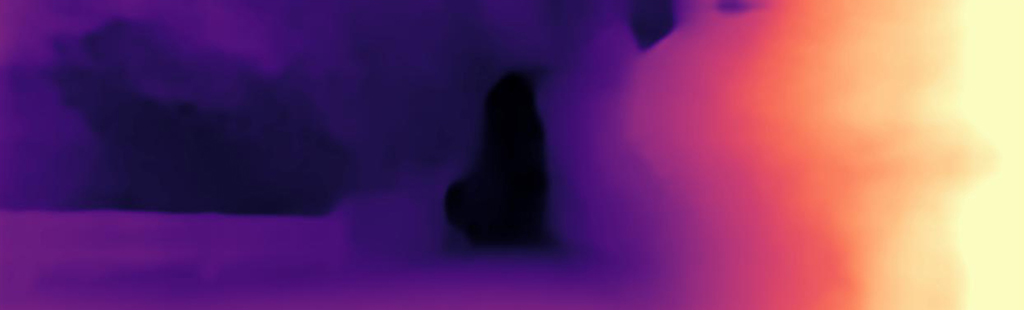} \\
    \rotatebox[origin=l]{90}{\scalebox{0.6}[0.7]{\hspace{4mm}ChiTransformer}} \hspace{1pt}
    & \includegraphics[width = 0.35\textwidth]{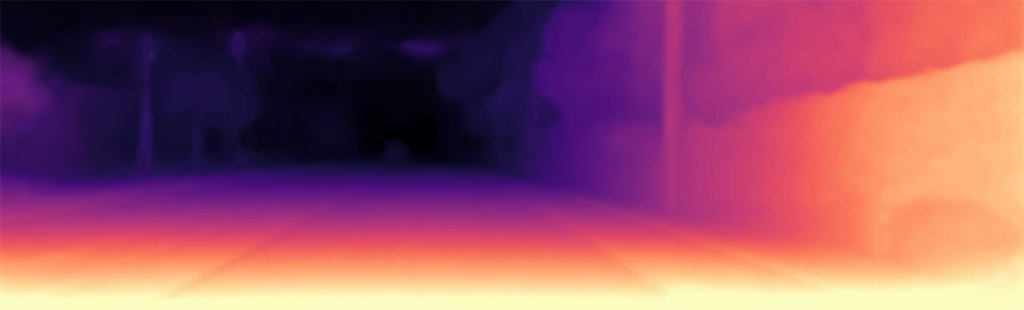} &
    \includegraphics[width = 0.35\textwidth]{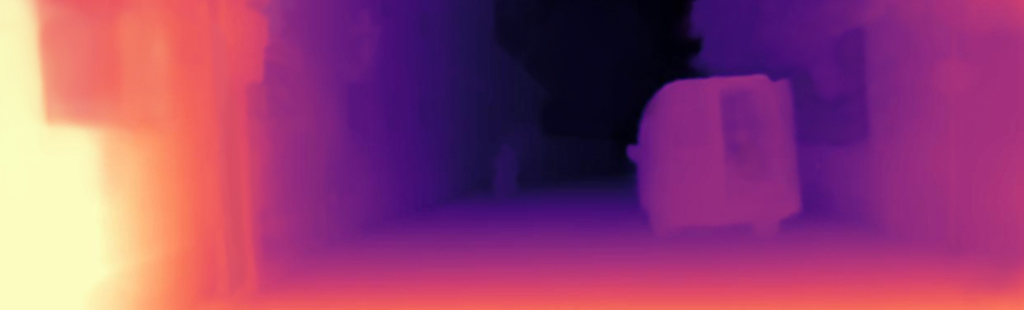} & \includegraphics[width = 0.35\textwidth]{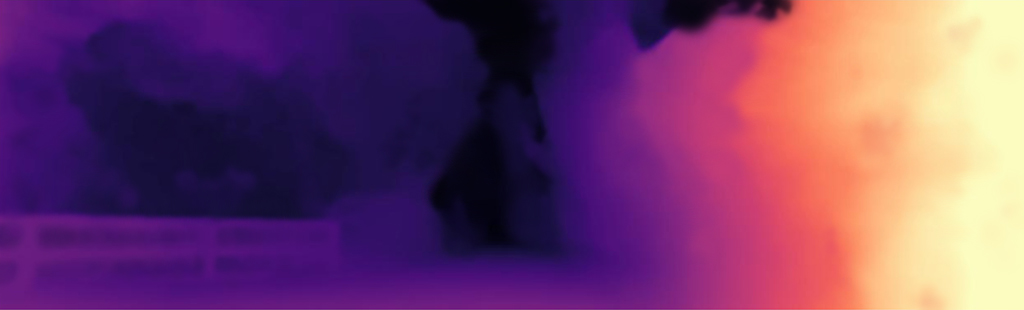} \\
    \end{tabular}
    \end{center}
\end{minipage}
\vspace{-2mm}\caption{Sample results compared with self-stereo-supervised fully-convolutional network Monodepth2. ChiTransformer shows better global coherence (e.g., sky region, sides of image) and provides feature consistent details.}
%\vspace{-3mm}
\label{figure:monovschi}
\end{figure*}

\vspace{3mm}
\section{Experiments}
\label{exprmt.}
The model is trained on KITTI 2015~\cite{kitti}. We show that our model significantly improves accuracy compared to its top CNN-based counterparts. Side-by-side comparisons are given in this section with the state-of-the-art self-supervised stereo methods~\cite{wang2020parallax,li2018occlusion,wang2021pvstereo}. Ablation study is conducted to validate that several features in ChiTransformer contribute to the improved prediction. Finally, we extend our model to fisheye images and yield visually satisfactory result.

\subsection{KITTI 2015 Eigen Split}

We divide the KITTI dataset following the method of Eigen et al.~\cite{eigen2014depth}. Same intrinsic parameters are applied to all images by setting the camera principal point at the image center and the focal length as the average focal length of KITTI. For stereo training, the relative pose of a stereo pair is set to be pure horizontal translation of a fixed length (0.54m) according to the KITTI sensor setup. For a fair comparison, depth is truncated to 80m according to standard practice~\cite{godard2017unsupervised}. 

\begin{figure}
\begin{minipage}{\columnwidth}
    \setlength{\tabcolsep}{0pt}
    \renewcommand{\arraystretch}{0.1}
\begin{center}
\scriptsize
\begin{tabularx}{\columnwidth}{@{}YYYYY@{}}

\multicolumn{5}{c}{ ~~~ \boldmagenta{---}focus region ~~~ \boldyellow{---}occluded area}\\
\multicolumn{5}{c}{\includegraphics[width=\linewidth]{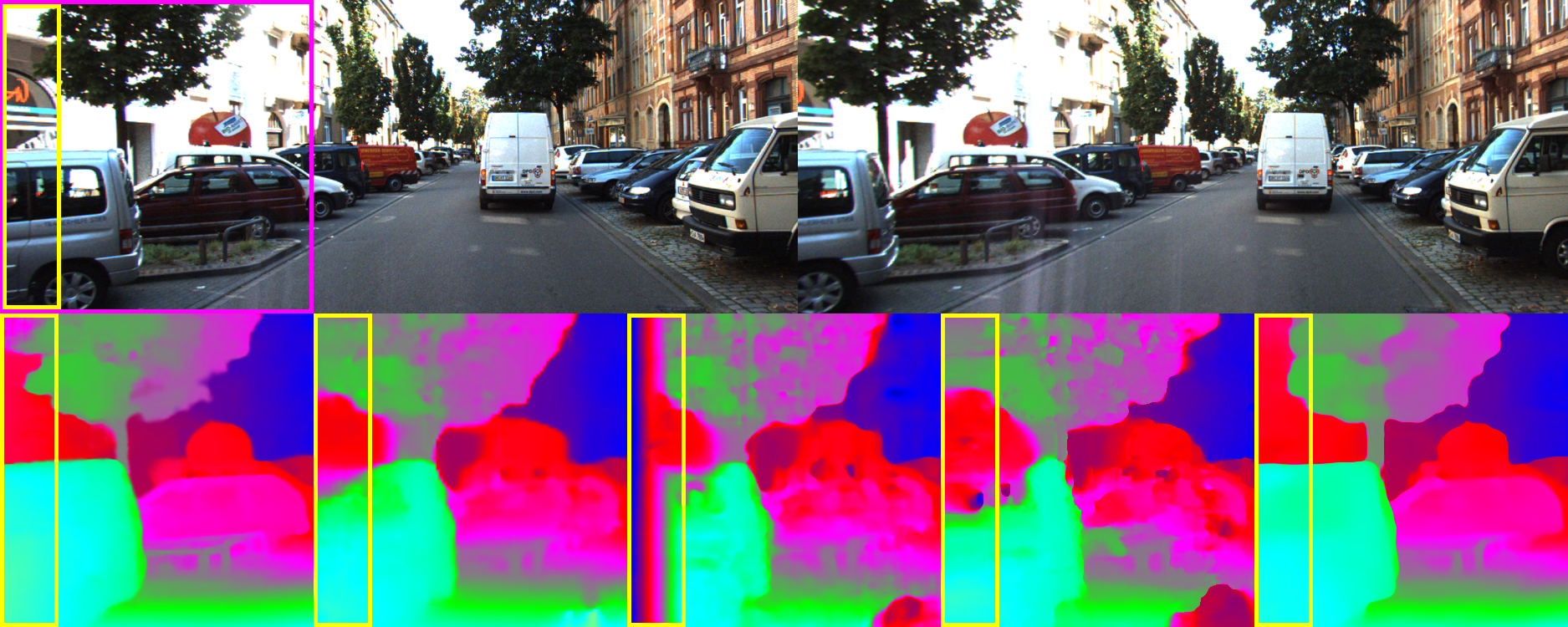}}\\
ChiTransformer & F2S\cite{liu2020flow2stereo} & OASM\cite{li2018occlusion} & PASM\cite{wang2020parallax} & PVS\cite{wang2021pvstereo}
\end{tabularx}
\end{center}
\end{minipage}
\vspace{-2mm}
\caption{Comparison of predictions in left side occluded area. }
\vspace{-5mm}
\label{figure:occlude_pred}
\end{figure}

\subsection{Quantitative Results}
We compare the results of the two different configurations of our model with state-of-the-art self-supervised stereo approaches. ChiT-8 has 4 SA layers followed by 4 rectification blocks, while ChiT-12 has 6 SA layers and 6 rectification blocks. The results in Table~\ref{table:stereo} show that ChiTransformer outperforms most of the existing methods, particularly in the prediction of the foreground regions. This trait is as expected that foreground regions are more likely to be abounded by distinctive features that benefit depth cue rectification. Qualitative results in occluded region comparing with existing self-supervised stereo methods are shown in Figure~\ref{figure:occlude_pred}. Their complete predictions are given in supplementary material. With depth cues from both views, ChiTransformer provides more details consistent to the image features compared to existing self-supervised methods.  

We also compare our method with top self-stereo-supervised MDE methods to show the reliability gain in terms of accuracy improvement. For a fair comparison, we choose the models that are trained on KITTI with stereo supervision. Methods trained over multiple datasets are not considered. Quantitative results are shown in Table~\ref{table:mono}. Side-by-side prediction comparison is shown in Figure~\ref{figure:monovschi}.
\vspace{2mm}

\begin{table*}
\centering
\begin{threeparttable}
\footnotesize
\caption{\vspace{-3mm}\textsc{Comparison with Self-stereo-supervised Monocular Methods}}
\label{table:mono}
\vspace{-3mm}
\begin{tabular}{m{7cm}cccccccl}
\thickhline
      Method      &AbsRel  &SqRel  &RMSE   &RMSE log   &$\delta<1.25$   &$\delta<1.25^2$     &$\delta<1.25^3$ \\
\hline
Garg et al.~\cite{garg2016unsupervised}   & 0.152   & 1.226   & 5.849   & 0.246   & 0.784  & 0.921  &0.967\\
3Net(R50)~\cite{poggi2018learning}     & 0.129   & 0.996   & 5.281   & 0.223   & 0.831  & 0.939  &0.974 \\
3Net(VGG)~\cite{poggi2018learning}     & 0.119   & 1.201   & 5.888   &0.208    & 0.844  & 0.941  &0.978 \\
SuperDepth+pp (1024$\times$382)~\cite{pillai2019superdepth}   & 0.112   & 0.875   & 4.968  & 0.207  & 0.852   &0.947  &0.977 \\
Monodepth2~\cite{godard2019digging}      & 0.109   &0.873   & 4.960   & 0.209   &0.864   &0.948   & 0.975 \\
\hline
\textbf{ChiT-12} (ours)   & \textbf{0.073}   & \textbf{0.634}   & \textbf{3.105}   & \textbf{0.118}   & \textbf{0.924}  & \textbf{0.989}  &\textbf{0.997} \\
\thickhline
\end{tabular}
\end{threeparttable}
\begin{minipage}{\linewidth}\footnotesize
    \vspace{1mm}
     All models listed in the table above are trained with self-supervised methods using stereo pair. Same as the monocular methods, ChiTransformer relies on depth cues to estimate depth, only with extra information from a second image.
\end{minipage}
\vspace{-15pt}
\end{table*}

\begin{table}
\begin{threeparttable}
\scriptsize
\caption{\textsc{Ablation Study}}
\label{table:ablation}
\begin{tabularx}{\columnwidth}{l p{6mm}p{6mm}p{6mm}p{7mm}p{7mm}p{7mm}}

\thickhline
                            &\scalebox{.9}[1.0]{AbsRel} 
                            &\scalebox{.9}[1.0]{RMSE} 
                            &\scalebox{.9}[1.0]{RMSElog} 
                            &\scalebox{.9}[1.0]{$\delta\!<\!1.25$} &\scalebox{.9}[1.0]{$\delta\!<\!1.25^2$} &\scalebox{.9}[1.0]{$\delta\!<\!1.25^3$}\\
                            
\hspace{-1mm}ChiT+P         &0.106 & 4.845 & 0.204& 0.878 & 0.960 & 0.981\\
\hspace{-1mm}ChiT+G+LEG     &0.101 &4.783  &0.203  &0.895  &0.966 &0.983\\
\hspace{-1mm}ChiT+P+Linear  &0.092 & 4.535 & 0.201 &0.889  &0.964 &0.987\\
\hspace{-1mm}ChiT+P+LEG     &0.085 & 3.924 & 0.181 & 0.906 &0.979 &0.991\\

\thickhline
\end{tabularx}
\begin{tablenotes}
    \footnotesize
      \item \hspace{-.8mm}Evaluations for different settings of ChiTransformer (ChiT) trained on KITTI 2015 with Eigen split. "P" denotes the polarized attention. "G" stands for the direct learning of matrix $G$. "LEG" represents the feature of learnable epipolar geometry. "Linear" is the single line attention zone. ChiT with only P enabled has the lowest score due to inferior token separability. With the addition of LEG, the model: ChiT+P+LEG, becomes the top performer and show the advantage of P over G compared with ChiT+G+LEG. ChiT+P+Linear has the 2nd best performance. The serration effect due to "Linear" is largely mitigated by the long range context information and SA layers. 
    \end{tablenotes}
    \vspace{-3mm}
\end{threeparttable}
\end{table}

\renewcommand{\arraystretch}{1}

\vspace{-10pt}
\subsection{Ablation Study}\vspace{-5pt}
To understand how each major feature contributes to the overall performance of ChiTransformer, ablation study is conducted by suppressing or activating specific components of the model. We observe that each component in our model is designed to push the performance a bit forward which aggregates into a sizable improvement. Here we provide some insights on the major features based on observation.
\newline\textbf{Self-attention layer} largely improves the separability of each token with long range complex contextual information. Without SA layer, the retrieval process would take up a hopping behavior and result in erroneous predictions.
\newline\textbf{Polarized attention} We learn the matrix $G$ through its spectral decomposition to gain more control over its behavior. Direct learning of $G$ tends to result in feature negligence as the major features contained within a token dominate or take all the reward. With the complementary feature highlighting-suppressing strategy as we desire $\prod{\Lambda}$ to be close to $\mathbf{I}$, the features from both parties can be attended. Meanwhile, since $\Lambda$ is not porous with zeros, i.e., no Lasso regularization involved, all information contained in the token is more or less attended.
\newline\textbf{Learnable Epipolar Geometry}
Intuitively, for rectified stereo a pixel pair is guaranteed to reside in the same horizontal line and hence the attending space should be that line. However, the slotted attention region hurts the inter-line connection and cause serrated effects on vertical features even that feature is distinctive, e.g., an edge. Whilst the learnable epipolar geometry in GPCA solves the problem by allowing global but focused view over the lines and at the same time further improves the cross-line separability. Quantitative results are given in Table~\ref{table:ablation}. Qualitative results are given in supplementary material.

\begin{figure}
\begin{minipage}{\columnwidth}
    \renewcommand{\arraystretch}{0.1}
    \begin{center}
    \scriptsize
    \begin{tabular}{@{}c@{}c}
     \hspace{1pt}
    \includegraphics[width = 0.48\columnwidth]{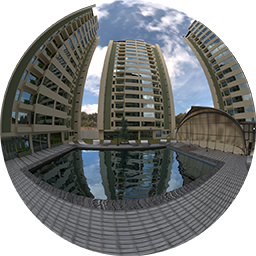}
    &\includegraphics[width = 0.48\columnwidth]{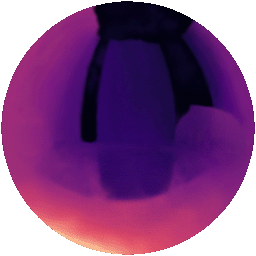}\vspace{2mm}\\
    Synthetic fisheye image & Planar depth estimation
    \end{tabular}
    \end{center}
\end{minipage}
\vspace{-3mm}\caption{Example result of ChiTransformer for fisheye depth estimation. With learnable epipolar curve $\bs{v}_{pos, ij} = (1,a,b,c,d,e)_{ij}$ (constant term is set to 1 to avoid proportional scaling) and circular masks, ChiTransformer can directly work on circular image without warping.}
\vspace{-3mm}
\label{figure:fisheye}
\end{figure}

\vspace{-5pt}
\subsection{Example Results on Fisheye Images}\vspace{-5pt}
Finally, due to its versatility, ChiTransformer can be applied to non-rectilinear (e.g., fisheye) images without warping. An example result is provided in Figure~\ref{figure:fisheye}.

\vspace{-5pt}
\section{Conclusion}\label{conc}\vspace{-5pt}

By investigating the limitations of the two prevalent methodologies of depth estimation, we present ChiTransformer, a novel and versatile stereo model that generates reliable depth estimation with rectified depth cues instead of stereo matching. With the three major contributions: (1) polarized attention mechanism, (2) learnable epipolar geometry, and (3) the depth cue rectification method, our model outperforms the existing self-supervised stereo methods and achieves state-of-the-art accuracy. In addition, due to its versatility, ChiTransformer can be applied to fisheye images without warping, yielding visually satisfactory results.

%The model is trained with reprojection-constrained self-supervised technique. 

\vspace{-5pt}
\section{Acknowledgment}\vspace{-5pt}
This research was sponsored in part by VMware Inc. for its university research fund and the Army Research Laboratory under Cooperative Agreement \#W911NF-22-2-0025. The views and conclusions contained in this document are those of the authors and should not be interpreted as representing the official policies, either expressed or implied, of the Army Research Laboratory or the U.S. Government. The U.S. Government is authorized to reproduce and distribute reprints for Government purposes notwithstanding any copyright notation herein.

%%%%%%%%% REFERENCES
\newpage
{\small
\bibliographystyle{ieee_fullname}
\bibliography{egbib}
}

\end{document}